\documentclass[11pt]{article}

\usepackage{amsmath,amsthm, amssymb}
\usepackage[all]{xy}
\usepackage{graphicx}

\usepackage{url}
\usepackage{braket}
\usepackage{amssymb}

\title{ Info-Evo: Using Information Geometry to Guide Evolutionary Program Learning }
\author{Ben Goertzel  }

\begin{document}

\maketitle

\begin{abstract}
A novel optimization strategy, {\bf Info-Evo}, is described, in which natural gradient search using nonparametric Fisher information is used to provide
ongoing guidance to an evolutionary learning algorithm, so that the evolutionary process preferentially moves in the directions identified as "shortest paths" 
according to the natural gradient.   Some specifics regarding the application of this approach to automated program learning are reviewed, including a strategy for integrating Info-Evo into the MOSES program learning framework.
\end{abstract}

\section{Introduction}

The core strength of evolutionary learning is the wild, creative, general-purpose generativity of the evolutionary process.   The core weakness  of evolutionary learning is its tendency to spend a lot of time exploring dead ends, even in cases where a bit of analytical or problem-specific reasoning would be able to identify the dead-end as such.   

Given this situation, it is natural that researchers have explored ways of injecting analytical (in particular, probabilistic) inference into the core of evolutionary algorithms -- yielding a class of algorithms known as EDAs or Estimation of Distribution Algorithms \cite{Pelikan2002}.   EDAs have proved successful for many types of problems.  However, there is not yet a truly convincing EDA for optimizing problems centrally involving floating-point (rather than discrete) variables.   And attempts to use EDAs for automated program learning, while interesting, have also failed to yield dramatically successful results.

What is suggested here is an alternate high-level strategy for combining the creative generativity of evolutionary learning with the more precisely targeted learning enabled via probabilistic inference: To wit, the close coupling of an evolutionary learning algorithm with a probabilistic learning algorithm, so that the latter extrapolates from the learning of the former in real time, and provides information biasing and guiding the ongoing search of the former.  In other words, this strategy involves a learning loop of the form:

\begin{enumerate}
\item Given the population of current guesses, probabilistic inference makes suggestions regarding what {\it kinds} of guesses are likely to be more successful than current ones
\item Evolutionary learning is used to vary the current population to find new guesses, in a way that is biased according to the suggestions made by probabilistic inference
\item The new guesses are evaluated and inserted into the population; unless a sufficiently good answer has been found or the allotted time has run out, we return to Step 1 and keep going
\end{enumerate}

This high-level strategy could be instantiated in many different ways; e.g. in \cite{BBM} it has been proposed to hybridize the MOSES program learning algorithm \cite{Looks2006} and the Probabilistic Logic Networks inference system \cite{PLN} in such a fashion.    Here we propose a particular instantiation called {\bf Info-Evo}, in which the role of the probabilistic inference algorithm is filled by information-geometric ("natural gradient") search on the space of "promise landscapes" -- a particular space of probability distributions over genotype space.

The relationship between information geometry and evolutionary learning has been highlighted in recent work by \cite{Ludovic2011} and others, which demonstrates that some successful learning algorithms like CMA-ES and PBIL may be modeled mathematically as natural gradient search, under appropriate assumptions.   One conceptual message of this work is that: Some algorithms that seem complex when viewed in terms of search over a certain space $S$, appear much simpler when viewed in terms of search over the space of probability distributions over $S$.  However, so far there has emerged no tractable way to use this perspective to derive search algorithms taking sophisticated account of dependencies within genotypes, in the manner of BOA \cite{Pelikan2005} and other advanced Estimation of Distribution algorithms.

Info-Evo integrates information-geometric and evolutionary  methods in a related but different way, via using the former to provide ongoing guidance to the latter, in a tightly integrated fashion.   The goal is to retain the exploratory generativity of evolutionary learning, but to reduce the inefficiency typical of evolutionary learning by using information geometry to estimate promising directions for search, and guide evolution along these directions.  This is enabled on the technical level by ideas from \cite{Mio2005} which explain how to do natural gradient search effectively in spaces of nonparametric probability distributions, and by the introduction of the novel framework of promise landscapes.  A promise landscape, roughly, is a fitness landscape for an optimization problem,  that is re-weighted to assign greater fitness to those candidate solutions that seem more promising as initial points for ongoing search.

To discuss Info-Evo precisely, it is useful to distinguish the base {\bf scoring function}, which is the objective function or initial fitness function one is looking to maximize; from the {\bf fitness function} which may include the scoring function along with other aspects.   For instance, in the current version of the MOSES evolutionary program learning algorithm \cite{Looks2007}, the fitness function includes the scoring function, plus penalty terms aimed to penalize overly complex programs, and encourage diversity among the programs learned.   Info-Evo can work with complexity and diversity penalties, but also adds additional terms to the fitness function, intended to guide search in directions estimated as likely to be valuable.

A rough verbal description of the Info-Evo process runs as follows:

\begin{enumerate}
\item Enumerate the population $S_n$ of $n$ programs whose scores have already been evaluated (by setting fitness equal to the base scoring function $\zeta$)
\item Heuristically estimate a promise landscape  $\Pi_\zeta$ for this population
\item Consider the promise landscape $\Pi_\zeta$ as a point in the $n-1$ dimensional space consisting of all possible promise landscapes defined over the same population of $n$ programs (with a metric defined from an inner product based on $\Pi_\zeta$)
\item Find geodesic rays in this $n-1$ dimensional space, beginning from $\Pi_\zeta$
\item Advance a distance $\gamma$ along each of these geodesic rays, thus obtaining a new promise distribution from each ray
\item For each of these new promise distributions $\Pi^i_\zeta$ (or only for the best few), create a new population of $m$ programs
\item In each of these new populations, use the corresponding promise distribution  $\Pi^i_\zeta$ to create a fitness function to guide an evolutionary search algorithm (which may use mutation, crossover, as well as Estimation of Distribution Algorithm type heuristics)
\begin{itemize}
\item During this evolution: For each new candidate program produced during the evolutionary search, estimate its fitness via the values of $\Pi^i_\zeta$  at its $k$ nearest neighbors within $S_n$ (where the proximity may be judged based on program structure, and/or based on program behavior).  If the estimated fitness is too low, don't evaluate the candidate
\end{itemize}
\item Add the newly evaluated programs to the population, and return to Step 1
\end{enumerate}

\noindent The terms and concepts in the above description will be elaborated in the following sections.

As is obvious from the above rough description, Info-Evo involves a fairly large amount of "machinery" that must be exercised in the course of the learning process, and therefore  is interesting in practice only in cases where fitness evaluation is very costly.   Of course, a wide variety of highly important learning problems fall into this category.

The Info-Evo approach is fairly general, and will initially be outlined here as such.  However, we will also describe    in more depth the specifics of integration of Info-Evo with the MOSES automated programming learning approach \cite{Looks2006}, which is the context in which the development of Info-Evo occurred.   MOSES is a probabilistic evolutionary learning system, that carries out evolution in a number of distinct subpopulations called "demes", each of which works in a different subset of program space, utilizing different input features and a distinct program structure.   Integrating Info-Evo into MOSES involves adding an additional layer, in which each deme spawns sets of temporary sub-demes corresponding to different promise distributions, as in Step 6 above.   We believe that MOSES with integrated Info-Evo will constitute a more powerful learning framework than the current MOSES version, especially for cases such as programs centrally reliant on floating-point inputs, and programs with complex control structures.   Integrating Info-Evo into MOSES will also provide a software basis for integration of other probabilistic learning methods such as PLN into MOSES.

\section{The Info-Evo Algorithm}

In this section the basic idea of the Info-Evo algorithm is described intuitively, in more detail than the above rough outline.  The following section provides further mathematical explication for the more intricate portions.

The first key concept to be introduced is that of a {\bf promise landscape}.   The "fitness landscape" corresponding to the problem of maximizing function $\zeta$ over space $S$, is generally defined as the graph of $\zeta$ over $S$.   A promise landscape is a related but not identical notion.   The {\bf promise function} $\Pi_\zeta$ corresponding to the scoring (objective) function $\zeta$ and the collection of guesses $S_n = (s_1, \ldots, s_n), s_i \in S$, is defined as follows: $\Pi_\zeta(x)$ represents the estimated probability (based on the given body of evidence and/or a given set of heuristics) that searching in the vicinity of $x$ for $y$ so that  $\zeta(y) > max_i \zeta(s_i)$ is likely to yield value.   The promise landscape is the graph of the promise function.

As thus articulated, the promise function is a conceptual notion that could be fully formalized in various ways.   Given a specific search algorithm $\mathcal{A}$ and an amount $R$ of computational resources, one could define $\Pi_\zeta(y)$ as the expectation of $ \frac{\zeta(z)} {max_i \zeta{s_i} }$, where: $z$ is the value found by  $\mathcal{A}$ (using resources $R$ and beginning from initial value $y$) that yields the highest value of $\zeta$.   We will not rely on this or any other formalization of the notion of a promise function here, but will merely use the promise function as a heuristic device within a proposed learning algorithm.  However, to prove theorems about the effectiveness of the methods described here, a formalization of the notion of a promise function would likely be required.

One of the important ideas of Info-Evo is to use estimates of a promise function to guide evolutionary learning.  That is: instead of just using $\zeta$ as the fitness function for evolution, one uses the promise function $\Pi_\zeta$.  This is similar to the utilization of fitness scaling in evolutionary learning, however considerably more general that fitness scaling as commonly practiced.  A traditionally scaled fitness function may be, in some cases, a better approximation of the promise function than the original fitness function.   However, it may also be useful to look at promise functions that are not naturally thought of in terms of simply rescaling fitness values, but that are derived by more sophisticated transformations of fitness values.

Some simple heuristics may be valuable for estimating the promise function (as required in {\bf Step 2} in the algorithm outline from the previous section).   For example, for each $s_i \in S_n$, one can look at:

\begin{itemize}
\item $p_{LM, i}$ = the probability that $s_i$ is near a local maximum of $\zeta$
\item $p_{GM, i}$ = the probability that $s_i$ is near a global maximum of $\zeta$
\end{itemize}

\noindent These may be estimated as follows:

\begin{itemize}
\item  $p_{LM, i}$  may be heuristically estimated as higher , depending on the ratio $\frac { \zeta(s_i)}{ max_{r \in N_k(s_r)} \zeta(s_r) }$ , where $N_k(s_r)$ consists of the $k$ nearest neighbors of $s_i$ among the elements of $S_n$,

 \item $p_{GM, i}$  may be heuristically estimated as higher, depending on the ratio $\frac{ \zeta(s_i)}{ max_r \zeta(s_r) }$
\end{itemize}

\noindent Given reasonable estimates of these quantities, once can heuristically define a promise function via

$$
\Pi_\zeta(s_i) = w_\zeta \zeta(s_i) + w_{LM} p_{LM, i}  + w_{GM} p_{GM, i}
$$
   
\noindent (Thus completing {\bf Step 2} of the algorithm outline.)  This formalizes the idea that it's not worth spending too much time hunting around previously evaluated points $s_i$ which, based on the available evidence, look likely to be far away from the global maximum, or worse yet, which don't even look to be in an upward direction relative to their neighbors.

However, this kind of heuristic estimate of the promise function is not yet the core of Info-Evo.   The central idea is, rather, to build promise functions embodying reasonable guesses about the overall directions that evolution should explore, based not just on localized analysis of the fitness landscape, but based on global study of the shortest paths (geodesics) through the space of promise landscapes.  Given a promise landscape (formed e.g. by the above heuristics), the space of "nearby" (in an appropriate abstract space) promise landscapes is constructed ({\bf Step 3} in the algorithm outline), and geodesic rays leading away from the initial promise landscape are determined ({\bf Step 4}).   New, candidate promise landscapes are then created, via following these geodesic rays a moderate distance from the initial promise landscape ({\bf Step 5}).   These candidate promise landscapes are then used to bias evolution of a new subpopulation ({\bf Steps 6, 7}), spawned from the previous subpopulation that was used to determine the initial promise landscape.   The ensuing evolutionary learning generates new fitness evaluations, which are used to spawn a new "initial" promise landscape, which is used to seed the geodesic-based process all over again.  

A little more formally: If the prior promise function for evolutionary learning over the space $S$ was $\Pi_\zeta$, the modified promise functions formed would be

$$
\Pi^j_\zeta(x) = h( \zeta(x) , \omega(x, p^j ) )  
$$

\noindent where 

\begin{itemize}
\item $h$ is monotone increasing in both arguments, and outputs nonnegative reals
\item The $p^j$ are probability distributions over $S_n$, representing estimated "right directions" relative to maximizing $\zeta$, obtained from the geodesic-based analysis mentioned above
\item $\omega( x, p^j) $ is a measure of the extent to which $x$ is in the direction of $p^j$ 
\end{itemize}

In general, many different inference algorithms (general purpose or problem domain specific) could be used to identify the "right direction" for evolution to proceed in.     However, in an Info-Evo context, the idea is to use a variation of natural gradient search.   One performs approximate natural gradient ascent on the fitness function $\Pi_\zeta$ in the space $\mathcal{P}$ of probability distributions over $S_n$ ({\bf Steps 4 and 5}), to be elaborated in the following section).  

There is no direct way to sample new genotypes from these new distributions $p^j$ , as they are defined as new weightings over the set of genotypes whose fitness has already been evaluated.   However, one can use such a distribution as a filter or guide for evolutionary learning ({\bf  Step 7}).   For instance, one can define $\omega( x, p^j)$ as the total probability, according to distribution $p^j$,  of the $k$ nearest neighbors of $x$ (where the neighbors are drawn from among the genotypes already evaluated).   The metric underlying the nearest neighbor calculation will be problem-specific, though in many cases it may be effectively taken as information geometry based as well.   Note that there are thus two separate metrics involved in Info-Evo:

\begin{itemize}
\item the Fisher metric on the space of promise distributions, which is used to find the geodesic rays that are followed to create new promise distributions ({\bf Step 4})
\item the metric on the space of individual programs $s_i$ (used  as part of fitness evaluation and estimation in {\bf Step 7}), which may be based on program structure and/ or program behavior; and may in some cases also be a Fisher metric if one chooses to represent a program as a probability distribution in some fashion
\end{itemize}

It will also be valuable to use the nearest neighbors of a new candidate genotype for fitness estimation, in order to determine which genotypes get their fitness evaluated in the first place.    In this case, one rarely bothers to evaluate a genotype that doesn't have reasonable similarity to the genotypes estimated to be in the right direction.    Then, once one does evaluate a genotype, its fitness is boosted according to its similarity to other genotypes estimated to be in the right direction.  The evaluation of new genotypes leads to a new space of samples $S_n$, over which the next set of probability distributions for use in the algorithm may be defined.

Given the assumption that $S$ is a program space, one has two kinds of similarity to play with: {\bf genotypic} (similarity between program trees, or other syntactic representations of programs) and {\bf phenotypic} (similarity between program {\it behaviors}, e.g. execution traces; or in the case of programs doing symbolic regression, lists of input/output pairs).    Use of syntactic similarity makes sense to the extent that syntax-semantics correlation holds for the program representation and fitness function question \cite{Looks2006}.   It will generally be valuable to use both types of similarity in defining the neighbors to use for fitness estimation, and the optimal balance between these is bound to be somewhat domain-specific, i.e. dependent on both the particular scoring function and the particular input types and internal program operations in use.

The fitness estimation phase is one place where Info-Evo explicitly accounts for probabilistic dependencies among portions of the genotypes, in the manner of BOA \cite{Pelikan2005} and other Estimation of Distribution algorithms.    Suppose, for instance, that the combination of having an AND in position 3 in a program tree, together with having an OR in position 5 of the program tree, tends to make a program tree fit (for the specific fitness function under consideration).    Let us suppose that one of the "right direction" distributions found reflects this, thus providing a higher weight to programs that possess this combination internally.   Then, programs containing this combination will have a smaller syntactic distance to the programs weighted highly by the right-direction distribution, and hence will be less likely to get filtered out prior to fitness evaluation.

(Probabilistic dependencies may also be used in Info-Evo via an explicit inclusion of EDA operations in the underlying evolutionary algorithm.   And, the search for geodesics implicitly takes into account for dependencies as well.)

\section{Geodesics in Promise Function Space}

We now elaborate further on the portion of Info-Evo involving finding geodesics in the space of promise functions ({\bf Steps 3,4,5}).  This is the most mathematically intricate part of the Info-Evo process, though in the end it boils down to some relatively straightforward algorithms.

As above, suppose we have a population $S_n$ of $n$ programs $s_1, ... , s_n$, each of which has had its fitness evaluated according to fitness function $\zeta$.   We may then set up an information-geometric framework for distributions over $S_n$, following the example of what \cite{Mio2005}  does for distributions over a line segment.

A probability distribution $p$ over the program space $S_n$, and its log-likelihood $\Phi(x) = \log p(x) $, may then be defined in terms of the vectors $( p(s_1), \ldots, p(s_n) )$ and $( \Phi(s_1), \ldots, \Phi(s_n) )$.   For instance, we can define a probability distribution $p_n^{\zeta}$ on $S_n$ via assigning each $s_i$ a probability $p_n^{\zeta}(s_i)$ proportional to its fitness value $\zeta(s_i)$.   This reflects the heuristic that the maximizers of $\zeta$ are more likely to lie near the samples that provide higher values of $\zeta$.  Similarly, we can set up a distribution for a promise function as defined above, $p_n^{\Pi_\zeta}$.  The promise function distribution can be viewed as a version of the fitness function distribution, with weight shifted away from those genotypes $s_i$ that appear unpromising as centers of neighborhoods for ongoing search, and toward other genotypes $s_i$ that appear more promising.

At each $\Phi \in \mathbb{R}^n$, we may consider the inner product

$$
\Braket{  f, g } = \sum_{i=1,\ldots,n} ^n  f_i g_i e^{\Phi_i}
$$

\noindent for $f, g \in \mathbb{R}^n$.    If $\Phi$ represents a promise distribution, this inner product gives higher weight to coordinate values corresponding to promising vicinities for search.  The inner product of $f$ and $g$ will be large if, roughly speaking, $f$ and $g$ both give high weight to programs that are considered to have high promise.   If $f$ and $g$ give high weight to many of the same programs, but all these programs tend to have low promise, the inner product will not be large.

If $ \sum_{i=1,\ldots,n} ^n e^{\Phi_i} =1 $ then $\Phi$ represents a probability distribution over $S_n$.   

Where

$$
F(\Phi) = \sum_{i=1,\ldots,n} ^n e^{\Phi_i} 
$$

\noindent we may define the differential of $F$ at $\Phi$, evaluated at $f \in \mathbb{R}^n$, as 

$$
dF_{\Phi}(f) = \sum_{i=1,\ldots,n} ^n  f_i e^{\Phi_i} = \Braket { f, 1}_ {\Phi}
$$

\noindent From this we may observe that, with respect to the inner product $\Braket { , }_\Phi $, the gradient of $F$ at $\Phi$ is

$$
\nabla F (\Phi) = (1,...,1)
$$

\noindent As the gradient of a function is orthogonal to its level sets, this implies that the level sets of $F$ are all orthogonal to $(1,...,1)$, independently of $\Phi$.   This implies that $ \mathcal{P}_n \equiv F^{-1}(1)$, the space of vectors $\Phi \in \mathbb{R}^n$ representing log-probability distributions over $S_n$, is an ($n-1$ dimensional) submanifold of $\mathbb{R}^n$.  

The tangent space $T_\Phi \mathcal{P}_n$ consists of all vectors $f \in \mathbb{R}^n$ satisfying $\Braket {f, 1}_\Phi =0$, i.e. so that $f$ is orthogonal to $(1,\ldots, 1)$ with respect to $\Braket { , }_\Phi $.    I.e., it consists of all functions with zero expectation with respect to the inner product $\Braket { , }_\Phi $.  In the case where $\Phi$ is defined as a promise distribution, this means all functions with zero expected log-promise. 

The {\it geodesic distance} between $\Psi, \Phi \in \mathcal{P}_n$ will be denoted $d(\Phi, \Psi)$.  This measures how far it is from one function to another, proceeding along the shortest path through function space, according to the Fisher-Rao metric; which means that locally at each point along the path, distance is defined consistently with the metric in the inner product defined above.   For instance, at $\Phi$ distance is measured consistently with $\Braket { , }_\Phi $; at  $\Psi$ distance is measured consistently with $\Braket { , }_\Psi $; and similarly at the intermediary points along the path between $\Phi$ and $\Psi$.

A core idea of Info-Evo is, given $S_n$, to look at the space $ \mathcal{P}_n $, and perform search in this space via beginning from $\Phi_n^{\Pi_\zeta}$  and proceeding along geodesics from this point according to a certain step size (and when a choice must be made, preferentially along geodesics in directions that appear most likely to lead to large values of $\Pi_\zeta$).  Given a new distribution in $ \mathcal{P}_n $, found by this method, one evaluates it by using it to bias the fitness function for evolution of a population of guesses attempting to maximize $\Pi_\zeta$, such as outlined above.   The new fitness evaluations done during this process enable construction of a new, larger space of evaluated programs $S_n$, on which a new space $ \mathcal{P}_n $ of distributions may be constructed.

\cite{Mio2005} proposes a shooting method for finding the geodesic rays in $ \mathcal{P}_n $ beginning from an initial point (in the set-up here, $\Phi_n^{\Pi_\zeta}$) .   However, shooting methods can be unreliable, as pointed out in \cite{Eberly2008}, especially where (as in the present case) a good guess for the initial direction is not at hand.   Instead, for the present application, we suggest that an implementation of Dijkstra's algorithm with a dynamically created grid, as described in \cite{Eberly2008}, would likely be more appropriate.   If this proves inadequately precise or too costly, one could use Dijkstra's algorithm on a relatively coarse grid to get a sense of some endpoints likely to lie near geodesic rays passing through $\Phi_n^{\Pi_\zeta}$ , and then use the hierarchical method described in \cite{Eberly2008} to find approximately geodesic polylines between $\Phi_n^{\Pi_\zeta}$  and these endpoints \footnote{code for this hierarchical method is given at \url{http://www.geometrictools.com/LibMathematics/CurvesSurfacesVolumes/CurvesSurfacesVolumes.html} }.

To recap {\bf Steps 4 and 5} of Info-Evo, then: Given approximations of $k$ geodesic rays $r_1,\ldots, r_k$ passing through $\Phi_n^{\Pi_\zeta}$ , one can then generate $k$ points $\Phi_n^{1}, \ldots, \Phi_n^{k}$  in $ \mathcal{P}_n $, lying distance $\gamma$ (a step size parameter) from $\Phi_n^{\Pi_\zeta}$ along each of the rays.   One can then create a modified  promise function $\Pi_{\zeta,i}$ corresponding to each of these points, i.e.

$$
\Pi_{\zeta,i}(x) = h( \zeta(x) , \omega(x,max_i \Phi_n^{i} ) )  
$$

\noindent where

\begin{itemize}
\item $h$ is monotone increasing in both arguments, and outputs nonnegative reals, e.g. $h(x,y) = x*y$
\item $\omega(x,y) $ is the magnitude of the projection of $x$ on the line between $\Phi_n^{\Pi_\zeta}$ and $y$
\end{itemize}

\noindent Each of these modified promise functions is used for  fitness evaluation and estimation, in a newly formed evolving population of guesses.   After a certain number of new fitness evaluations, the totality of programs evaluated is used to form a new space $S^m$, with $m>n$, and the process of determining geodesic directions is begun again.

\section{Embedding Info-Evo within MOSES}

The MOSES evolutionary program learning framework \cite{Looks2006} provides a natural context for Info-Evo to operate within.  MOSES carries out program learning within a large set of islands or "demes", all of which share the same fitness function, but each of which corresponds to a particular set of program inputs and a particular "exemplar" program used to initiate search in the space of programs taking subsets of these inputs.   To implement Info-Evo within MOSES, the most natural course is to associate each deme with a set of sub-demes, each of which carries out learning based on a particular promise function generated by learning in its parent deme.  According to a crude estimate, this increases the resource requirements of MOSES by a constant factor equal roughly to the number of sub-demes per deme (however, this estimate doesn't take into account the possibility that Info-Evo might result in faster learning, thus saving resources overall; or might decrease the number of demes needed, etc.).

The nature of the similarity measurement used for promise landscape base fitness estimation within Info-Evo, will of course depend on the particular subclass of programs being evolved and the domain in question.   However, one appealing option is to re-use the portions of the natural gradient search process from Info-Evo, and use estimates of the Fisher metric to measure distance between programs.    That is: given a set of programs and $v$ input vectors at which they have all been evaluated, one can represent each of the programs as an $v$-dimensional vector.   Given two programs, one can then calculate the distance between them according to the Fisher metric, using the polyline method as outlined in \cite{Eberly2008}.

\section{Discussion}

The algorithmic and mathematical particulars suggested above become somewhat involved, so it's worth reiterating what they're trying to accomplish conceptually.  The core goal is to combine the wild, creative generativity of evolutionary learning, with the greater focus and efficiency that come from probabilistic modeling.  EDAs attempt to do this, effectively, by replacing or augmenting evolutionary operations with probabilistic modeling operations.  This may well be a good idea in some cases.   But Info-Evo attempts to combine evolution and probability on  different level: via using probabilistic modeling on {\it population space} to provide high-level guidance to the evolutionary process.  This is not contradictory to the EDA approach; the different approaches can be used together.

The fact that CMA-ES and PBIL and other well-known algorithms can be viewed in terms of natural gradient ascent on population space (representing populations as probability distributions) is inspiring.   The approach taken here is a bit different in that Info-Evo doesn't try to find cheap approximations to the Fisher metric based on simplifying assumptions.  Rather, it somewhat brute-forces the Fisher metric calculations using Dijkstra's algorithm.   Further, rather than replacing evolutionary operations, it embraces them with all their low-level noise and waste, but seeks to direct them via using tweaked fitness functions representing estimated likely directions of evolution.    Info-Evo is thus focused on the case where the evaluation of the basic scoring measure underlying the fitness function is {\bf so} expensive that it's worthwhile doing a bunch of computational work to estimate likely best directions for evolution to proceed in.


\bibliographystyle{alpha}
\bibliography{bbm}

\end{document}